\documentclass{article}
\usepackage{spconf,amsmath,graphicx}
\usepackage{booktabs}
\usepackage{multirow}
\usepackage{url}
\usepackage{float}

\let\OLDthebibliography\thebibliography
\renewcommand\thebibliography[1]{
 \OLDthebibliography{#1}
 \setlength{\parskip}{3pt}
 \setlength{\itemsep}{3pt plus 0.2ex}
}

\title{Enhancing Indonesian Automatic Speech Recognition: \\ Evaluating Multilingual Models with Diverse Speech Variabilities}

\name{
 \begin{tabular}{c}
Aulia Adila\textsuperscript{1}\textsuperscript{*}\thanks{*This work was conducted while the first author was doing internship at HA3CI Laboratory, JAIST, Japan under JST Sakura Science Program.}, Dessi Lestari\textsuperscript{1}, Ayu Purwarianti\textsuperscript{1}, Dipta Tanaya\textsuperscript{2},\\ Kurniawati Azizah\textsuperscript{2}, Sakriani Sakti\textsuperscript{3,4}
 \end{tabular}
}

\address{
\begin{tabular}{c}
\textsuperscript{1}Institut Teknologi Bandung, Indonesia\\ 
\textsuperscript{2}University of Indonesia, Indonesia \\
\textsuperscript{3}Japan Advanced Institute of Science and Technology, Japan\\
\textsuperscript{4}Nara Institute of Science and Technology, Japan\\
\texttt{13519100@std.stei.itb.ac.id}, \\\texttt{\{dessipuji,ayu\}@itb.ac.id}, \\
\texttt{\{diptatanaya,kurniawati.azizah\}@cs.ui.ac.id}, \\\texttt{ssakti@is.naist.jp}
\end{tabular}
}

\begin{document}
\maketitle

\begin{abstract}
An ideal speech recognition model has the capability to transcribe speech accurately under various characteristics of speech signals, such as speaking style (read and spontaneous), speech context (formal and informal), and background noise conditions (clean and moderate). Building such a model requires a significant amount of training data with diverse speech characteristics. Currently, Indonesian data is dominated by read, formal, and clean speech, leading to a scarcity of Indonesian data with other speech variabilities. To develop Indonesian automatic speech recognition (ASR), we present our research on state-of-the-art speech recognition models, namely Massively Multilingual Speech (MMS) and Whisper, as well as compiling a dataset comprising Indonesian speech with variabilities to facilitate our study. We further investigate the models' predictive ability to transcribe Indonesian speech data across different variability groups. The best results were achieved by the Whisper fine-tuned model across datasets with various characteristics, as indicated by the decrease in word error rate (WER) and character error rate (CER). Moreover, we found that speaking style variability affected model performance the most.
\end{abstract}

\section{Introduction}
Automatic Speech Recognition (ASR) is a technology that converts spoken language into written text using algorithms \cite{yu_automatic_2015}. It is ideal for recognizing and transcribing accurate text under various conditions and sound signals. However, real-world applications face challenges due to variations in speech data, including differences in formal and informal contexts, speaking styles (read and spontaneous), and background noise conditions (clean and moderate).

Building an ASR model capable of handling these diverse speech characteristics requires a substantial amount of training data. In the context of the Indonesian language, however, the availability of large and diverse datasets for training is limited, which can impact the quality of a model if it is trained from scratch. To address this limitation, the use of transfer learning has emerged as a promising solution in the development of machine learning models \cite{kunze_transfer_2017}.

Transfer learning involves leveraging pre-existing, pre-trained models that already have a foundational understanding of related tasks. In our approach, we used fine-tuning, a method where a pre-trained model is further trained (or fine-tuned) on a new dataset. Recently, the development of Indonesian ASR has focused on pre-trained multilingual models for both the Indonesian language \cite{arisaputra_indonesian_2022,whisper_pratama_2024} and Indonesian ethnic languages \cite{Sakti2023LeveragingTM}, particularly with an emphasis on read and formal speech. 

Recognizing the scarcity of Indonesian data with diverse speech variations, our study aims to evaluate Indonesian ASR performance across different speech characteristic datasets. The models are built using a fine-tuning approach on state-of-the-art pre-trained multilingual models: Massively Multilingual Speech (MMS) \cite{pratap_scaling_2023} and Whisper \cite{radford_robust_2022}. To facilitate the study, we built a comprehensive Indonesian speech dataset, the IDSV Dataset, collected from several repositories to ensure that the diversity of speech characteristics is captured. Furthermore, we grouped the testing data based on speech variability categories for a more holistic evaluation.

\begin{table*}[t]
\centering
\caption{IDSV Dataset: Indonesian Speech Variabilities Dataset}
\resizebox{\textwidth}{!}{%
\begin{tabular}{@{}lcccccccccc@{}}
\toprule
\multicolumn{1}{c}{\multirow{2}{*}{\textbf{Dataset}}} & \multirow{2}{*}{\textbf{Domain}} & \multirow{2}{*}{\begin{tabular}[c]{@{}c@{}}\textbf{Speaking}\\ \textbf{Style}\end{tabular}} & \multirow{2}{*}{\textbf{Hours}} & \multirow{2}{*}{\begin{tabular}[c]{@{}c@{}}\textbf{Unique Spk.} \\ {\textbf{[}}\textbf{Overlap. Spk.}{\textbf{]}}\end{tabular}} & \multirow{2}{*}{\textbf{Utts.}} & \multirow{2}{*}{\textbf{Words}} & \multirow{2}{*}{\begin{tabular}[c]{@{}c@{}}\textbf{Utts. len.}\\ \textbf{(avg. words)}\end{tabular}} & \multicolumn{3}{c}{\textbf{Hours}} \\ \cmidrule(l){9-11} 
\multicolumn{1}{c}{}                         &                         &                                                                        &                        &                                                                                         &                        &                        &                                                                                    & \textbf{Train}   & \textbf{Dev}    & \textbf{Test}   \\ \midrule
News                                         & News                    & Read                                                                   & 20.38                  & 385 {[}0{]}                                                                             & 20,494                 & 97,732                 & 4.71                                                                               & 12.86   & 2.61   & 4.91   \\
Teldialog                                    & Phone dialogue          & Read                                                                   & 17.09                  & 388 {[}0{]}                                                                             & 18,917                 & 84,397                 & 4.44                                                                               & 10.73   & 2.26   & 4.10   \\
CV 13                                        & Open domain             & Read                                                                   & 15.75                  & 347 {[}0{]}                                                                             & 11,971                 & 89,676                 & 7.16                                                                               & 10.94   & 2.16   & 2.65   \\
FLEURS                                       & Wikipedia               & Read                                                                   & 9.28                   & 2,738 {[}0{]}                                                                           & 2,738                  & 51,091                 & 18.62                                                                              & 5.68    & 1.61   & 1.99   \\
DPR                                          & Parliament              & Spontan.                                                               & 5.67                   & 90 {[}0{]}                                                                              & 2,942                  & 44,516                 & 19.23                                                                              & 3.11    & 0.63   & 1.93   \\
KPK                                          & Parliament              & Spontan.                                                               & 2.53                   & 51 {[}0{]}                                                                              & 3,589                  & 28,738                 & 8.69                                                                               & 1.83    & 0.56   & 0.70   \\
KY                                           & Parliament              & Spontan.                                                               & 4.98                   & 50 {[}0{]}                                                                              & 7,364                  & 41,507                 & 5.94                                                                               & 3.16    & 0.89   & 0.93   \\
Talkshow                                     & YouTube                 & Spontan.                                                               & 1.11                   & 34 {[}0{]}                                                                              & 611                    & 7,273                  & 11.33                                                                              & 0.56    & 0.41   & 0.14   \\
Podcast-1                                    & YouTube                 & Spontan.                                                               & 3.28                   & 10 {[}0{]}                                                                              & 5,609                  & 39,934                 & 7.03                                                                               & 2.42    & 0.08   & 0.78   \\
Podcast-2                                    & YouTube                 & Spontan.                                                               & 0.042                  & 3 {[}2{]}                                                                               & 39                     & 572                    & 19.76                                                                              & 0.04    & -      & 0.002  \\
Podcast-3                                    & YouTube                 & Spontan.                                                               & 0.43                   & 9 {[}4{]}                                                                               & 220                    & 4,684                  & 21.29                                                                              & -       & -      & 0.43   \\ \bottomrule
\end{tabular}%
}
\label{tab:dataset}
\end{table*}

We first provide an overview of relevant research on Indonesian ASR and other speech recognition frameworks. Sections 3 through 5 describe the datasets, multilingual models, and experimental pipeline used. Section 6 presents the performance comparison results, and the final section offers our conclusions.


\vspace{-2mm}
\section{RELEVANT WORKS}
Initial Indonesian speech recognition was developed in a phoneme-based system using the cross-language approach \cite{sakti_rapid_2005}. It adapted the acoustic information from the English language to obtain an Indonesian acoustic model with a full phoneme set. Another work has developed a large vocabulary continuous speech recognition (LVCSR) system for the Indonesian language in a classical approach, i.e., building the acoustic modeling, language modeling, and lexicon \cite{sakti_development_2008}. The speech data used in both studies consisted of read, formal, and clean speech.

Over the past decade, all-neural ASR architecture have emerged, offering integrated, fully neural models that learn more reliably from data and require less prior experience. Researchers have employed pre-trained multilingual models, such as XLSR-53 \cite{arisaputra_indonesian_2022} and Whisper \cite{whisper_pratama_2024}, for Indonesian language and several pre-trained models to enhance speech recognition performance in Indonesian ethnic languages \cite{Sakti2023LeveragingTM}, focusing on read and formal speech data. However, the ability to handle other speech variabilities has not yet been explored.


Recent advancements in ASR research have introduced state-of-the-art end-to-end models like MMS \cite{pratap_scaling_2023} and Whisper \cite{radford_robust_2022}, which enhance speech recognition performance while reducing the need for extensive training data. The MMS model supports over 1,100 languages by utilizing publicly available religious texts and leveraging the wav2vec 2.0 framework \cite{wav2vec2} with self-supervised learning. It is fine-tuned for tasks such as speech recognition, speech synthesis, and language identification. The Whisper model, on the other hand, uses a large amount of labeled audio data in a weakly-supervised manner without fine-tuning, and it can recognize speech, identify languages, detect voice activity, and translate speech across more than 96 languages. Despite their capabilities, neither model has been explored for handling variabilities in Indonesian speech data.


\section{DATASET}


To handle the variability of speech data represented by different characteristics, the dataset used in this study consists of Indonesian speech data compiled from several existing repositories. Table \ref{tab:dataset} summarizes our IDSV Dataset\footnote{IDSV Dataset: Indonesian Speech Variabilities Dataset}.

Parts of our dataset were taken from existing Indonesian speech datasets \cite{sakti_development_2008, sakti_2013} obtained in previous research projects \cite{sakti_2004, SAKTI2013509, Sakti2014RecentPI}. These include read speech recorded from news script reading and phone dialogue, referred to as LVCSR News and LVCSR Teldialog, respectively. We also utilized commonly used multilingual speech datasets from the Common Voice Corpus \cite{common_voice}, specifically version 13.0, and FLEURS \cite{fleurs}, both focusing on the Indonesian language. Additionally, we collected speech datasets from Indonesian parliament meetings (DPR, KPK, KY) and Indonesian online talk shows and podcasts from YouTube (Talkshow, Podcast-1, Podcast-2, Podcast-3) \cite{prosa}. We noticed that some utterances from Podcast-2 and Podcast-3 contained overlapping speech from multiple speakers. Since these datasets had the least amount of data, we included them only in the development or testing sets. Overall, our IDSV Dataset contains approximately 80.54 hours of speech, divided into training, development, and testing sets, with a train+dev to test ratio of 77\% to 23\%.



Subsequently, the data was processed through several steps, which includes standardizing metadata, cleaning ground truth transcripts, and dividing the dataset into training, development, and testing sets. Repeated recordings were removed to ensure fairness and accuracy. Lastly, an analysis was conducted on the testing data to categorize speech based on various characteristics.

\begin{table}[h]
\caption{IDSV test dataset groupings based on the speech variabilities}
\resizebox{\columnwidth}{!}{%
\begin{tabular}{@{}cccclc@{}}
\toprule
\textbf{Group} & \textbf{\begin{tabular}[c]{@{}c@{}}Speaking\\ Style\end{tabular}} & \textbf{\begin{tabular}[c]{@{}c@{}}Speech\\ Context\end{tabular}} & \textbf{\begin{tabular}[c]{@{}c@{}}Background\\ Noise\end{tabular}} & \multicolumn{1}{c}{\textbf{\begin{tabular}[c]{@{}c@{}}SNR\\ (dB)\end{tabular}}} & \textbf{Dataset} \\ \midrule
RFC & Read & Formal & Clean & \textgreater 44 & \begin{tabular}[c]{@{}c@{}}News\\ Teldialog\end{tabular} \\ \midrule
RFM & Read & Formal & Moderate & 17-19 & \begin{tabular}[c]{@{}c@{}}CV 13\\ FLEURS\end{tabular} \\ \midrule
SFM & Spontaneous & Formal & Moderate & 12-16 & \begin{tabular}[c]{@{}c@{}}DPR\\ KPK\\ KY\end{tabular} \\ \midrule
SIC & Spontaneous & Informal & Clean & 30-43 & \begin{tabular}[c]{@{}c@{}}Podcast-1\\ Podcast-3\end{tabular} \\ \midrule
SIM & Spontaneous & Informal & Moderate & 19-21 & \begin{tabular}[c]{@{}c@{}}Talkshow\\ Podcast-2\end{tabular} \\ \bottomrule
\end{tabular}%
}

\label{tab:test-dataset}
\end{table}

The data was grouped based on the type of conversation and word choice, categorizing the recordings as either formal or informal. Speaking style was classified according to how sentences in the speech dataset were formed: read from written text (read) or spoken naturally and spontaneously (spontaneous). For background noise conditions, classifications were based on the signal-to-noise ratio (SNR) \cite{papic_signal--noise_2010} and recording location, differentiating between studio recordings with minimal noise (high SNR, above 30 dB) and non-studio settings with significant background noise (low SNR, below 30 dB). As a result, we divided the test data into several groups, as shown in Table \ref{tab:test-dataset}.

\vspace{-1.5mm}
\section{MULTILINGUAL SPEECH MODELS}


In this study, we investigated the MMS and Whisper models to develop Indonesian speech recognition models capable of transcribing speech with various characteristics. We used fine-tuning, which involved further training the pre-trained models on a specific dataset and task. In this case, we used our IDSV dataset for the downstream ASR task.

Due to computational resource limitations, we fine-tuned the \texttt{mms-300m} model with 317 million parameters and the \texttt{whisper-small} model with 244 million parameters. We selected these models because they have a similar number of parameters, allowing for a fair comparison.

\subsection{The Massively Multilingual Speech (MMS)}

MMS is built upon pre-trained self-supervised speech representation model (wav2vec 2.0) and uses both labeled and unlabeled datasets. The labeled dataset (MMS-lab) contains speech-text pairs for 1,107 languages, obtained from New Testament online text and audio recorded in a single-speaker setting. The unlabeled dataset (MMS-unlab) is sourced from religious recordings and songs, resulting in 7.7K hours of data covering 3,809 languages. 

The unlabeled datasets alongside with other open-source corpora (491K hours; 1,406 languages) were used to pre-train the model, resulting pre-trained model serves as a baseline for several downstream tasks, including ASR, language identification, and speech synthesis, available in two model sizes: \texttt{mms-300m} (0.3 billion parameters) and \texttt{mms-1b} (1 billion parameters). For the speech recognition task, a linear layer is added to map outputs to transcriptions, and the model is fine-tuned using the Connectionist Temporal Classification (CTC) criterion \cite{ctc}  on the MMS-lab dataset. A second fine-tuning step is performed for the \texttt{mms-1b} model, incorporating language-specific adapters. The final ASR model supports transcription in 1,107 languages, demonstrating the effectiveness of the MMS framework in multilingual speech recognition.

\begin{table*}[t]
\centering
\caption{WER and CER score of ASR models evaluation on IDSV test dataset across speech variabilities groups}
\begin{tabular}{@{}lllllllllllll@{}}
\toprule
 & \multicolumn{2}{c}{\textbf{ALL}} & \multicolumn{2}{c}{\textbf{RFC}} & \multicolumn{2}{c}{\textbf{RFM}} & \multicolumn{2}{c}{\textbf{SFM}} & \multicolumn{2}{c}{\textbf{SIC}} & \multicolumn{2}{c}{\textbf{SIM}} \\ \cmidrule(l){2-13} 
 & WER & CER & WER & CER & WER & CER & WER & CER & WER & CER & WER & CER \\ \midrule
\textit{Open-source ASR} &  &  &  &  &  &  &  &  &  &  &  &  \\
mms-1b-all & 29.30 & 9.57 & 23.47 & 5.71 & 15.53 & \textbf{3.45} & 37.93 & 15.12 & 54.29 & 22.41 & 35.89 & 12.01 \\
whisper-small & 30.87 & 13.84 & 27.22 & 9.85 & 23.04 & 7.36 & 38.85 & 20.64 & 40.66 & 25.41 & 22.85 & 11.56 \\ \midrule
\textit{Indonesian ASR (Ours)} &  &  &  &  &  &  &  &  &  &  &  &  \\
FT-mms (no LM) & 29.66 & 7.86 & 20.45 & 4.23 & 25.28 & 5.77 & 35.93 & 10.14 & 52.68 & 19.19 & 32.84 & 9.58 \\
FT-mms (with LM) & 19.95 & 6.45 & 13.75 & 3.24 & 15.64 & 4.31 & 23.36 & 8.14 & 39.75 & 17.84 & 22.25 & 8.19 \\
FT-whisper & \textbf{14.85} & \textbf{5.44} & \textbf{10.38} & \textbf{2.71} & \textbf{13.12} & 3.95 & \textbf{15.98} & \textbf{6.78} & \textbf{29.64} & \textbf{14.81} & \textbf{13.04} & \textbf{5.85} \\ \bottomrule
\end{tabular}%
\label{tab:results}
\end{table*}


\subsection{Whisper}

Whisper is a speech recognition model trained on 680k hours of labeled audio data, covering over 96 languages. It eliminates the need for dataset-specific fine-tuning for ASR tasks and uses an extensive text normalizer to reduce penalization for non-semantic differences. The training dataset was constructed from audio-text pairs found on the Internet, covering various environments, recording setups, speakers, and languages. To improve transcript quality, automated filtering methods were developed, and machine-generated transcripts were removed from the training sets.

The model architecture is a Transformer-based encoder-decoder model, which uses learned position embeddings and tied input-output token representations. The model was trained on multilingual speech recognition, speech translation, language identification, and voice activity detection. The multitask training format uses a set of special tokens as task specifiers. The Whisper model family consists of several versions: \texttt{whisper-tiny} (39 million parameters), \texttt{whisper-base} (74 million parameters), \texttt{whisper-small} (244 million parameters), \texttt{whisper-\\medium} (769 million parameters), and \texttt{whisper-large} (1.5 billion parameters).

\section{EXPERIMENTAL SETTINGS}

\begin{figure}[t]
\begin{center}
\includegraphics[width=0.8\linewidth]{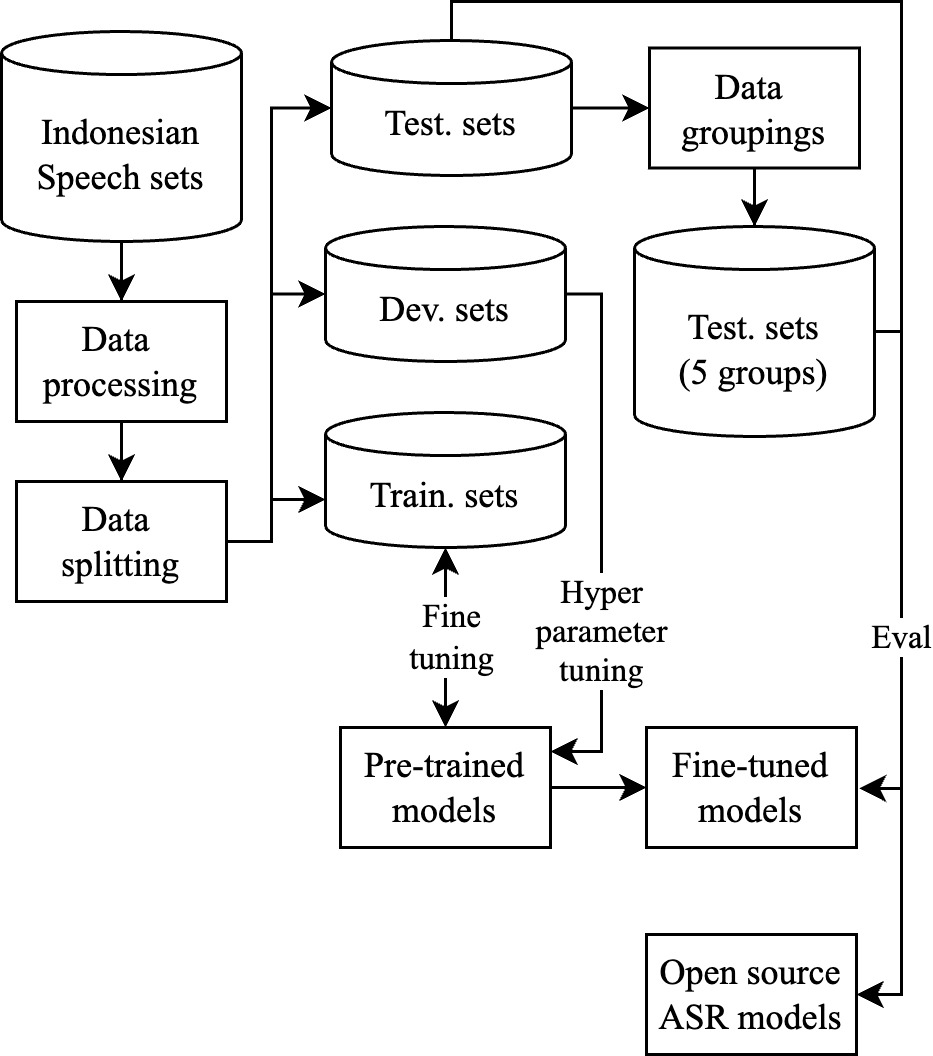}
\caption{Overview of our experiment pipeline}
\label{fig:pipeline}
\end{center}
\vspace{-5mm}
\end{figure}

We built the Indonesian ASR models by fine-tuning the \texttt{mms-300m} and \texttt{whisper-small} pre-trained models on our IDSV dataset, naming the resulting models \texttt{FT-mms} and \texttt{FT-whisper}, respectively. To assess each model's capability in the speech recognition task, we evaluated our fine-tuned models on the evaluation sets summarized in Table \ref{tab:test-dataset}. Additionally, we evaluated the readily available ASR models \texttt{mms-1b-all} and \texttt{whisper-small} without fine-tuning to compare the results with those of our fine-tuned models. Prediction accuracy was measured using the Word Error Rate (WER) and Character Error Rate (CER) evaluation metrics. A decrease in WER indicates an improvement in the model's ability to understand word sequences, while a decrease in CER reflects better performance in predicting speech at the character level, such as recognizing phoneme sound units. The overall experiment pipeline is shown in Fig. \ref{fig:pipeline}.

We utilized Hugging Face \cite{huggingface} for fine-tuning our models for ASR tasks. To build the \texttt{FT-mms} model, we fine-tuned the \texttt{mms-300m} pre-trained model using the CTC criterion. We froze the CNN layers, which were sufficiently trained during the pre-training phase, to preserve their ability to extract acoustically meaningful but context-independent features from the raw speech signal. During fine-tuning, we used the Adam optimizer \cite{adam} with a learning rate of 1e-5. Training was limited to 30 epochs, with up to 4 of the best models being selected and saved during the process. Furthermore, we conducted another experiment by adding a 5-gram language model, trained from our training set vocabularies using KenLM \cite{heafield-2011-kenlm}, on top of the CTC layer. This provided a fairer comparison with the \texttt{FT-whisper} model, which uses a neural sequence model as a decoder, functioning as both a language model and trained on billions of web tokens.

Whisper, on the other hand, was pre-trained on a vast amount of labeled audio data, so it requires minimal additional fine-tuning to achieve high performance in ASR tasks \cite{radford_robust_2022}. Our \texttt{FT-whisper} model was built by fine-tuning the \texttt{whisper-small} model with a maximum of 10,000 steps and 100 warm-up steps, while our \texttt{FT-mms} model was built with a maximum of 91,680 training steps. During fine-tuning, we used the Adam optimizer \cite{adam} with a learning rate of 1.25e-5. We incorporated hyperparameter tuning using our development sets to optimize the performance of both models.

\section{EVALUATION RESULTS}
As shown in Table~\ref{tab:results}, almost all of our models outperformed the open-source ASR models due to fine-tuning, as expected. However, the MMS fine-tuned models showed an increase in both WER and CER on the RFM test sets compared to \texttt{mms-1b-all}. We considered this degraded performance is due to a decline in the model's ability to predict the RFM data, as this type of data is widely available and easy to obtain on open-source platforms, which likely dominate the model training data. Therefore, fine-tuning the model towards other speech variabilities reduced its initial ability to predict RFM speech characteristics. Further comparison of our MMS fine-tuned models showed that the addition of the language model improved performance across all test set groups. This improvement can be attributed to the language model's ability to better capture linguistic context, leading to more accurate transcription results.

However, when compared to our Whisper fine-tuned model, \texttt{FT-whisper} outperformed almost all other models in predicting speech variabilities. The fine-tuning process also significantly increased performance compared to the open-source model, with a decrease of $16.02$ in WER and $8.4$ in CER across all test sets. We also observed very competitive results between the open-source models, \texttt{whisper-small} and \texttt{mms-1b-all}, which can be attributed to the differences in parameter sizes between the models.

By further analyzing each speech variability group within the evaluated models, the RFM and RFC groups had the lowest WER and CER values among the test sets for the MMS models. This indicates that the MMS models, both with and without fine-tuning, already have sufficient knowledge of these speech characteristics due to their dominance in the training sets. Meanwhile, the results for Whisper show the lowest WER and CER in the SIM test group, demonstrating the model's robustness due to its ability to transcribe speech with characteristics found in real-world conditions, rather than ideal lab conditions.

There is a group of test data that obtained the highest WER and CER across all evaluated models, namely SIC. The best performance reached WER and CER values of $29.64$ and $14.81$, respectively, on our \texttt{FT-whisper} model, indicating unsatisfactory performance on the ASR downstream task. This is due to overlapping sentence recordings by different speakers in the test data, which were not sufficiently trained into the model, making the ASR task challenging to perform.

Thus, we would like to go upon a deeper investigation on the speech variabilities: speaking style, speech context, and background noise. As we examined further, speaking style has significantly influence the MMS model's predictive ability, as seen in both RFC and RFM sets results discrepancy comparing to SFM, SIC, and SIM. Meanwhile, as for the Whisper model, evaluation on RFC and RFM sets achieved the low WER and CER on the fine-tuned model, comparatively similar with SIM sets as it has obtained the highest performance on the open-source model. This means that the model has effectively learned the speech characteristics through fine-tuning phase. 

Regarding the speech context, the MMS models were analyzed to have a better ability in predicting the formal speech comparing to the informal speech,  due to its training sets certain speech characteristic dominance. However, the speaking style remains a greater influence on the model's accuracy. This can be seen from the groups of read data which have the best accuracy consistently. On the other hand, the open-source Whisper model has a competitive result on formal and informal test sets, means that this variability has less influence on the model's accuracy. Additionally, the noisy or clean background noise conditions are not significantly affect the models predictive ability. This can be observed from the inconsistency of the trend on the models without and with fine-tuning process in the RFC and RFM sets. While both models are already adept at predicting speech with moderate noise, Whisper has showcasing the superior ability.

As the prediction errors were further analyzed, we found that the models were still falsely predicting specific names, homophones, and terms due to the speaking dialect. Moreover, it was found that some recordings had wrongly chosen formal/informal words in their ground truth transcription. Code-switching with English has also become one issue found in the dataset.

\section{CONCLUSION}

This paper presents our contribution to enhancing Indonesian ASR in predicting speech variabilities by fine-tuning state-of-the-art multilingual speech models, MMS and Whisper. We also compiled several existing datasets to form a comprehensive Indonesian speech dataset representing speech variabilities, namely the IDSV Dataset. Furthermore, we evaluated the open-source MMS and Whisper models, as well as our fine-tuned models, on our test datasets, which were grouped based on three speech characteristics: speaking style, speech context, and background noise.

The experiment results show that fine-tuning improved the models' predictive ability, with our Whisper fine-tuned model, \texttt{FT-whisper}, achieving the best accuracy on test data groups: RFC, RFM, SFM, SIC, SIM, as well as on the whole test set in general, with the highest accuracy on the SIM sets. Furthermore, we found that speaking style variability had the most significant influence on model performance. On the other hand, all models showed unsatisfactory results on the SIC set due to the overlapping utterances spoken by multiple speakers, which were not sufficiently trained during the fine-tuning phase.

However, the analyzed speech variabilities could be extended beyond speech context, speaking style, and background noise conditions to include other aspects such as speech dialect/accent, speaker physiology, and the emotional state of the speaker. Additionally, it is recommended to perform fine-tuning on more advanced models by leveraging better computing resources.


\section{Acknowledgments}
Part of this work is supported by JSPS KAKENHI Grant Numbers JP21H05054 and JP23K21681, as well as JST Sakura Science Program.

\bibliographystyle{IEEEbib}
\bibliography{refs}

\end{document}